\documentclass[12pt]{iopart}

\usepackage[utf8]{inputenc}
\usepackage[T1]{fontenc}
\makeatletter
\expandafter\let\csname equation*\endcsname\@undefined
\expandafter\let\csname endequation*\endcsname\@undefined
\makeatother
\usepackage{amsmath}
\usepackage{amssymb}
\usepackage{graphicx}
\usepackage{xcolor}
\usepackage{siunitx}
\usepackage{tabularx}
\usepackage{xspace}
\usepackage{multirow}
\usepackage[hidelinks]{hyperref}

\hypersetup{
    colorlinks=false,
    linkcolor=blue,
    citecolor=green,
    filecolor=magenta,
    urlcolor=cyan
}

\newcommand{\fig}{Fig.~}
\newcommand{\eq}{Eq.~}
\newcommand{\myRef}{Ref.}

\begin{document}

\title{Spintronics for image classification: performance benchmarking via data-driven simulations}

\author{A.~Moureaux$^1$, C.~Chopin$^2$, S.~de~Wergifosse$^1$, L.~Jacques$^3$ and F.~Abreu~Araujo$^1$}

\address{$^1$ Institute of Condensed Matter and Nanosciences, Université catholique de Louvain, Place Croix du Sud 1, 1348 Louvain-la-Neuve, Belgium}
\address{$^2$ Univ. Grenoble Alpes, CEA, CNRS, Grenoble INP, SPINTEC, 3800 Grenoble, France}
\address{$^3$ Institute for Information and Communication Technologies, Electronics and Applied Mathematics, Université catholique de Louvain, Avenue Georges Lemaître 4, 1348 Louvain-la-Neuve, Belgium}

\ead{anatole.moureaux@uclouvain.be}

\begin{abstract}
    We present a demonstration of image classification using an extreme learning machine (ELM) based on a unique simulated magnetic tunnel junction (MTJ) delayed in time.
    As the ground state of the MTJ is a magnetic vortex, we refer to it as a vortex-based spin-torque oscillator (STVO).
    The dynamics of the magnetic vortex is simulated with a model called the data-driven Thiele equation approach (DD-TEA).
    This allows to avoid the constraints associated with repeated experimental manipulation for hyperparameters search and benchmarking.
    We showcase the versatility of our implementation by using it successfully for classification tasks on the MNIST, EMNIST-letters and Fashion MNIST datasets.
    Through simulations, we show that within an ELM with a sufficient number of parameters, the performance reached using the STVO dynamics as a source of nonlinearity is equivalent to the ones obtained with classical software activation functions such as the reLU and the sigmoid.
    While achieving state-of-the-art accuracy levels on the MNIST dataset, our model's performance on EMNIST-letters and Fashion MNIST is lower due to the simplicity of the network architecture and the increased complexity of the data.
    We expect that the DD-TEA framework will enable the exploration of deeper and more complex STVO-based architectures, ultimately leading to improved classification accuracy.
\end{abstract}

\maketitle

\section{Introduction} 

The impact of artificial intelligence (AI) on various industries and our daily lives is undeniable~\cite{lecun2015deep}.
However, the widespread use of AI has raised concerns about its energy consumption and environmental impact during resource-intensive training and large-scale inference phases~\cite{dhar2020carbon, patterson2021carbon}.
With the rapid development of large generative models, the electricity demand of AI is now growing fast enough to even match that of small countries~\cite{devries2023growing}.
Moreover, conventional computers, despite decades of miniaturization and optimization, are nearing their limits in terms of computing power~\cite{wilkes1995memory, frank2001device, waldrop2016chips}.

To face these challenges, new approaches in artificial intelligence have emerged, aiming to surpass the limitations of digital frameworks~\cite{markovic2020physics}.
Among them, reservoir computing bypasses the training of the network parameters by using a large recurrent network with random and fixed connections~\cite{jaeger2004harnessing}.
Within this framework, the input data are projected into a higher-dimensional space in which the categories become linearly separable.
The parameters of the readout layer are learned using a simple linear regression, which is computationally efficient and can be performed in a single shot over the whole training set.
This approach hence allows to find a global minimum of the loss function while saving time, energy and avoiding the vanishing gradient problem observed with traditional algorithms relying on gradient descent and error backpropagation.
While reservoir computing was initially developed for processing time-dependent data, it can be generalized to the classification of static data with extreme learning machines (ELM)~\cite{huang2006extreme}.
ELMs are feed-forward neural networks where a randomly initialized hidden layer stays fixed and acts as a random feature mapper.
The output layer is the only part of the network requiring training, which is usually achieved through a simple linear regression, similarly to reservoir computing.
The fixed nature of the reservoir (or of the hidden layer in the case of ELMs) makes it well-suited for hardware implementations, enabling the creation of dedicated hardware analog processors to optimize performance and energy efficiency~\cite{tanaka2019recent}.
Research has also demonstrated that such reservoirs could be implemented using a single nonlinear device delayed in time without sacrificing performance~\cite{appeltant2011information,ortin2015unified,borghi2021reservoir}, by trading on the execution time.
By sampling the output of the device repeatedly, one is able to emulate a large number of virtual nodes connected in time instead of space.
This method, called time-multiplexing, further eases the hardware implementation of these systems~\cite{borghi2021reservoir, larger2017high, riou2017neuromorphic, abreu2020role}.

\begin{figure}
    \centering
    \includegraphics[width=.5\columnwidth]{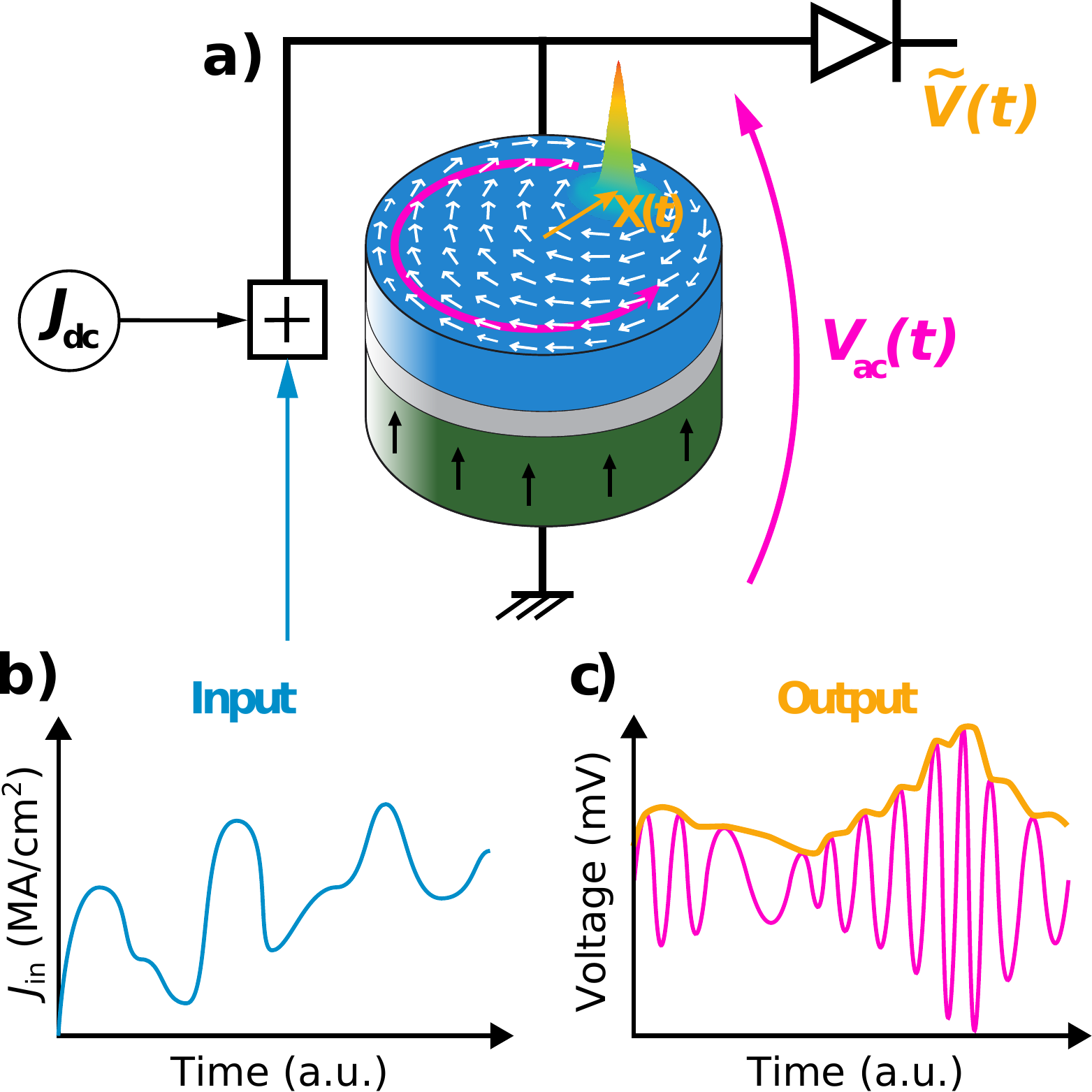}
    \caption{\textbf{(a)} Radio-frequency oscillations of the vortex core reduced position $s(t)$ are triggered by \textbf{(b)} the injection of an amplitude-modulated signal $J_\text{in}(t)$ added to a bias current density $J_\text{dc}$ in the STVO.\@
    \textbf{(c)} The non-linearity of $s(t)$ can be retrieved experimentally by recording the envelope $\tilde{V}(t)$ of the oscillating voltage $V_\text{ac}(t)$ across the oscillator (best seen in color).}\label{fig:stvo_in_out}
\end{figure}

Efficient data classification using physical reservoir computing has been demonstrated in various works utilizing spintronic oscillators known as vortex-based spin-torque oscillators~\cite{pribiag2007magnetic} (STVOs).
STVOs are magnetic tunnel junctions (MTJs) whose behavior can be described using spintronics, the study of the transport of spin in magnetic materials and nanostructures~\cite{gaididei2010magnetic}.
The ground state magnetization of STVOs is a magnetic vortex, composed of in-plane curling magnetization with an out-of-plane magnetization vortex core at the center (\fig\ref{fig:stvo_in_out}a). 
When a current density signal is injected into a STVO (\fig\ref{fig:stvo_in_out}b), the vortex core undergoes circular oscillations in the plane of the MTJ (pink circular arrow in \fig\ref{fig:stvo_in_out}a) due to an effect called the spin-transfer torque~\cite{pribiag2007magnetic, ralph2008spin, guslienko2014nonlinear} (STT). 
Thanks to another phenomenon known as tunnel magnetoresistance\cite{yuasa2007giant} (TMR), stable oscillations of the electrical resistance are obtained, as well as voltage oscillations in the MHz range across the MTJ (\fig\ref{fig:stvo_in_out}c).
The amplitude of these voltage oscillations is nonlinearly related to the amplitude of the input signal, making STVOs suitable for nonlinear data transformation in reservoir computing applications~\cite{abreu2022ampere}.
This nonlinearity can be retrieved from the evolution of an internal state quantity of the STVO: the time-varying reduced position of the vortex core $s(t)=\vert\vert X(t)\vert\vert/R$, where $\vert\vert X(t)\vert\vert$ is the time-dependent position of the vortex core in the plane of the MTJ and $R$ is the radius of the MTJ~\cite{araujo2022data}.
STVOs offer numerous advantages, such as low power consumption, minimal noise and compatibility with complementary metal-oxide-semiconductor (CMOS) technology~\cite{yogendra2015coupled}, making them attractive candidates for nonlinear data transformation in reservoir computing applications~\cite{torrejon2017neuromorphic, romera2018vowel, markovic2019reservoir,abreu2020role,grollier2020neuromorphic}.

Despite the progress in STVO-based hardware computing systems, a significant challenge lies in their development, optimization and testing at large scale.
Indeed, the performance of these systems is highly dependent on the choice of the hyperparameters, such as the number of virtual nodes in the reservoir (or the hidden layer in the case of ELMs), the bias current density $J_\text{dc}$, the peak-to-peak amplitude of the input signal, and the sampling rate.
Despite being crucial, hyperparameters search remains a time-consuming and resource-intensive task, especially within experimental frameworks.
To overcome these limitations, simulating STVOs and their complex dynamics proves to be the easiest approach.
Micromagnetic simulations~\cite{leliaert2019tomorrow} (MMS) offer highly accurate results but are time and energy-consuming, making them impractical for large scale simulations.
A computationally lighter alternative is the Thiele equation approach (TEA), which uses a single ordinary differential equation for each direction of space to model the STVO dynamics~\cite{thiele1973steady}.
TEA is unfortunately merely qualitative in the steady and transient oscillatory states, which respectively describe stable magnetization oscillations and transitions from a stable state to another one. 
Both of these regimes however play a key role in nonlinearly transforming the data~\cite{abreu2020role, araujo2022data}.
A recent solution to this challenge is the data-driven Thiele equation approach (DD-TEA) developed by Abreu Araujo \textit{et al.}~\cite{araujo2022data}.
This semi-analytical model accurately describes the STVO dynamics in both steady and transient states by using numerical parameters extracted from a small set of MMS.\@
DD-TEA enables STVO simulation with the same level of accuracy as MMS, but with a significant acceleration factor of $9$ orders of magnitude~\cite{moureaux2023neuromorphic}.
While the numerical parameters used in this work are extracted for a specific STVO geometry, the underlying methodology consisting in fitting the coefficients of a reduced analytical model to a small set of micromagnetic simulations is not tied to this particular device.
It can in principle be also transposed to other oscillator-based nanodevices whose dynamics admit an analytical description, making the present approach a template rather than a single-device solution.

The present study shows the ability of a simulated STVO-based ELM to achieve state-of-the-art accuracy for image classification on the MNIST dataset.
First, we present the implementation of a time-multiplexed ELM based on a single STVO delayed in time, as well as the data preprocessing and training procedures.
We then explain how the STVO dynamics is simulated using the DD-TEA framework.
Finally, we evaluate the network's performance in classifying various types of images, such as handwritten digits from MNIST~\cite{deng2012mnist}, latin characters from EMNIST-letters~\cite{cohen2017emnist} and clothing items from Fashion MNIST~\cite{xiao2017fashion} (FMNIST).
We compare the STVO dynamics to more conventional nonlinear functions (reLU and sigmoid) in order to highlight the usability of the vortex dynamics as a hardware activation function.
Those studies are carried out for various numbers of virtual nodes in the ELM to demonstrate the potential of DD-TEA for large-scale hyperparameters research.

The data-driven Thiele equation approach underlying our STVO model was introduced in \myRef~\cite{araujo2022data}, and a first proof of concept of its use for accelerated neuromorphic computing was reported in \myRef~\cite{moureaux2023neuromorphic}, with an early single demonstration on the MNIST dataset.
The present work is however self-contained and extends these earlier reports in both model fidelity and scope.
The STVO nonlinearity is computed here with the full high-precision form of the model (Eqs.~\ref{eq:hptea}--\ref{eq:n}), rather than a reduced approximation, ensuring quantitative agreement with micromagnetic simulations across the whole operating range.
More importantly, we move from a single demonstration to a systematic study: we benchmark the same single-STVO ELM on three distinct image datasets, compare the STVO dynamics with standard software activation functions to establish their equivalence, and take advantage of the speed of the model to map the performance against the reservoir size.
All the results reported below are produced within a single pipeline, so that the present study can be read and reproduced independently of any earlier work.

\section{Methods}

\subsection{Time-multiplexed ELM using a single time-delayed STVO}
ELMs first send the input data in a random higher-dimension space, before transforming the projected data nonlinearly.
If the dimension of the projection space is sufficiently high, the categories can be separated using a simple linear classifier, which is the only part of the model requiring training.
The present work presents a partially physical ELM for classifying the MNIST dataset, where the nonlinear transformation is performed by a single STVO delayed in time, similarly to the single chain time-delayed reservoir presented in \myRef~\cite{ortin2015unified}.

\subsubsection{Preprocessing}
We first apply a denoising and dimensionality reduction procedure to the input data using principal component analysis~\cite{ringner2008principal, karamizadeh2013overview} (PCA).
PCA extracts the directions (\textit{i.e.}, linear combinations of pixels) along which the variance of the images is the highest.
This allows us to reduce the dimension of the images from $784$ pixels to $44$ pixels combinations while still explaining $80\%$ of the variance in the training set.
First, each image is flattened into a column vector $\mathbf{x}$ of $784$ pixels whose value is the grayscale intensity normalized between $0$ and $1$.
Once the PCA components matrix $\mathbf{C}$ is computed based on the training data, the significant components are extracted from each input image $\mathbf{x}$ accordingly to \eq\ref{eq:pca} (see blue part in \fig\ref{fig:elm}).

\begin{equation}
    \mathbf{x}' = \mathbf{C}\mathbf{x}
    \label{eq:pca}
\end{equation}  

\subsubsection{Random higher-dimension projection}

Each sample $\mathbf{x}'$ is then projected into a space of dimension $N_\theta >> 44$ using a random matrix $\mathbf{M}$ whose elements are drawn from $U(-1, 1)$ (\eq\ref{eq:masking}). 
Note that $\mathbf{M}$ is defined only once and stays fixed for the whole training and inference phases.
More pragmatically, this operation corresponds to propagating the data into a hidden layer of $N_\theta$ nodes with random weights (orange part in \fig\ref{fig:elm}).

\begin{equation}
    \mathbf{x}'' = \mathbf{M}\mathbf{x'}
    \label{eq:masking}
\end{equation} 

\subsubsection{Physical nonlinear transformation}

The projected data samples $\mathbf{x}''$ are then nonlinearly transformed using the STVO dynamics, which effectively corresponds to the application of a neural activation function.
Each $\mathbf{x}''$ is first converted into a current density signal $\mathbf{J}$ by scaling each of its elements linearly into a predefined range. 
The STVO transformation of $\mathbf{J}$ is simulated by computing the reduced position of the vortex core $\mathbf{s}$ using $\mathbf{J}$ as an input signal (\eq\ref{eq:nonlinearity}) and the model described later in subsection~\ref{modeling} (central part in \fig\ref{fig:elm}).
The value $1$ is stacked on top of the vector to account for the bias term in the linear classification step (see next subsection). 

\begin{equation}
    \mathbf{s} = s(\mathbf{J}) = \begin{bmatrix}1\\s(J (x''_1))\\ \vdots \\s(J(x''_{N_\theta}))\end{bmatrix}
    \label{eq:nonlinearity}
\end{equation} 

\subsubsection{Training}

The STVO output $\mathbf{s}$ of all the samples in the training set are concatenated to form the matrix $\mathbf{S}_\text{train}$ (dimension $(N_\theta +1) \times 60000$).
In parallel, the target matrix $\mathbf{T}_\text{train}$ (dimension $10 \times 60000$) is formed by encoding the category of each training sample in a one-hot vector (zeros everywhere except for a one at the index of the correct class).
The weights matrix of the output layer $\mathbf{W}_\text{out}$ (dimension $10 \times (N_\theta +1)$), which must satisfy the first part of \eq\ref{eq:Wdef}, is computed with the Moore-Penrose pseudo-inverse~\cite{courrieu2008fast, barata2012moore, ben2003generalized} of $\mathbf{S}_\text{train}$ (second part of \eq\ref{eq:Wdef}).

\begin{equation}
    \mathbf{W}_\text{out}\mathbf{S}_\text{train} = \mathbf{T}_\text{train} \Leftrightarrow \mathbf{W}_\text{out} = \mathbf{T}_\text{train}\mathbf{S}_\text{train}^\dagger
    \label{eq:Wdef}
\end{equation}

The linear system from \eq\ref{eq:Wdef} is strongly overdetermined, as the number of virtual nodes remains far below the number of training samples ($N_\theta \ll 60{,}000$).
Therefore, the regular Moore-Penrose pseudo-inverse can be used rather than a ridge regression, without observing any overfitting (Sec.~\ref{sec:results}).
We still verified this assumption by evaluating the use of a ridge regularization parameter $\lambda$ on a validation set.
The optimal parameter is negligible and the resulting test accuracy is similar from that of the pseudo-inverse, confirming that regularization is not necessary in this case.

\subsubsection{Linear classification}

\begin{figure}[ht]
    \centering
    \includegraphics[width=\columnwidth]{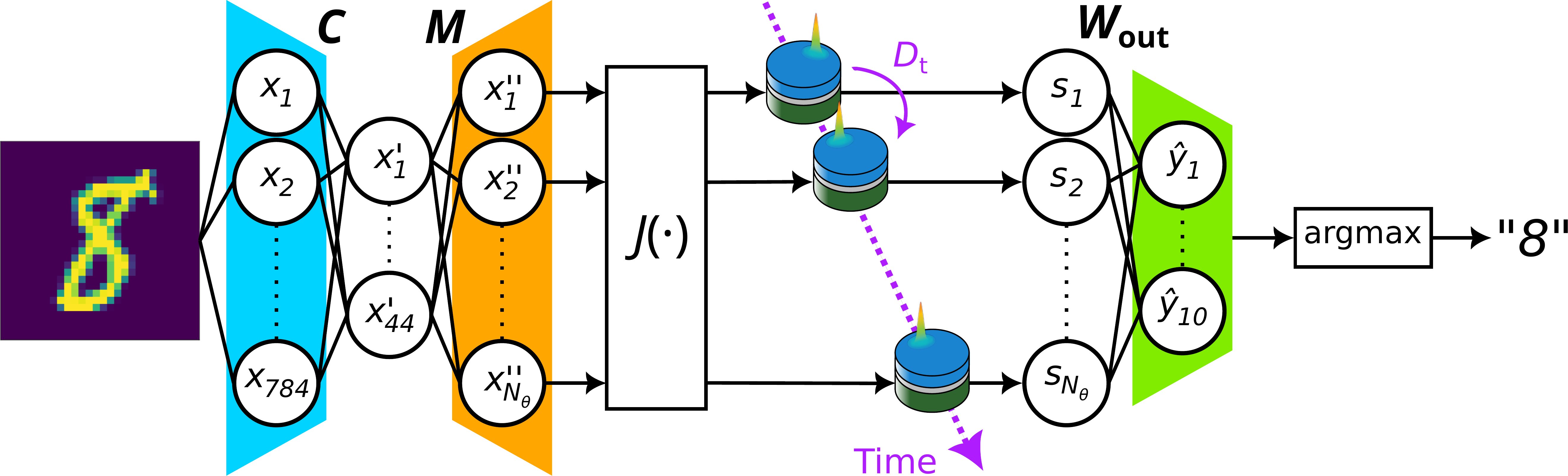}
    \caption{Partially physical ELM based on a single STVO delayed in time for image recognition.
    The dimensionality reduction, projection in a high-dimensional space, STVO transformation and linear classification steps are illustrated from left to right.}\label{fig:elm}
\end{figure}

As a result, $\mathbf{W}_\text{out}$ effectively contains the parameters of a linear mapping between a (new) STVO output $\mathbf{s}$ and the most likely one-hot categorical vector $\mathbf{\hat{y}}$ (\eq\ref{eq:output}).
The corresponding predicted digit $\hat{d}$ is obtained by selecting the index of the maximum element in $\mathbf{\hat{y}}$ using the argmax function (\eq\ref{eq:argmax}) as depicted in the rightmost part of \fig\ref{fig:elm}.

\begin{equation}
    \mathbf{\hat{y}} = \mathbf{W}_\text{out}\mathbf{s}
    \label{eq:output}
\end{equation} 

\begin{equation}
    \hat{d} = \text{argmax}(\mathbf{\hat{y}})
    \label{eq:argmax}
\end{equation} 

\subsection{Modeling the STVO dynamics}\label{modeling}

To compute \eq\ref{eq:nonlinearity} we simulate the reduced position $s$ of the vortex core for the input current density signal $\mathbf{J}$ made from the input data.
This can be analytically computed within the DD-TEA~\cite{araujo2022data} framework using the recurrent \eq\ref{eq:hptea}.  
The parameters $\alpha_i$, $\beta_i$ and $n_i$ are derived from the TEA and rule the time-dependent nonlinearity of the STVO dynamics.
In the DD-TEA framework, they are polynomials of the input signal $\mathbf{J}$ fitted through MMS~\cite{araujo2022data} (Eqs.~\ref{eq:alpha},~\ref{eq:beta} and~\ref{eq:n}). 
The input signal $\mathbf{J}$ is obtained by scaling the input data into a current density range of $\Delta J \approx 2.84$~MA/cm² centered around a bias density of $J_\text{dc} \approx 6.32$~MA/cm² (\textit{i.e.}, $\mathbf{J} \in [J_\text{dc}-\Delta J/2, J_\text{dc}+\Delta J/2]$), corresponding to a working current $I_\text{w} \approx 1.99$~mA and a signal voltage span of $\Delta V = 125$~mV across the oscillator.
The $D_\text{t}$ parameter represents the duration for which each value in $\mathbf{J}$ is injected into the STVO, and the sampling period of the output signal $\mathbf{s}$.
It is chosen shorter than the characteristic transient time of the STVO to ensure that the observed dynamics is always nonlinear, and is set to $D_\text{t} = 10$~ns in this work.
The initial value $s_0$ is the value of $s$ when $J_\text{dc}$ is injected into the oscillator for a duration $t >> D_\text{t}$. 
In the context of reservoir computing, the model described in \eq(\ref{eq:hptea}) plays the role of the internal nonlinear function implemented by the virtual STVOs in the reservoir~\cite{tanaka2019recent, larger2017high, abreu2020role, lukosevivcius2009reservoir, larger2012photonic, paquot2012optoelectronic}.

\begin{equation}
    s_i = s_{i-1}{\left[\left(1+\dfrac{s_{i-1}^{n_i}}{\alpha_{i}/\beta_i}\right)\exp\left(-n_i\alpha_i D_t\right)-\dfrac{s_{i-1}^{n_i}}{\alpha_i/\beta_i}\right]}^{-1/n_i}
    \label{eq:hptea}
\end{equation}

\begin{equation}
    \label{eq:alpha}
    \alpha_i = 6.64 J_{i} -39.97 
\end{equation}

\begin{equation}
    \label{eq:beta}
    \beta_i = -0.43 J_{i}-25.92
\end{equation}

\begin{align}
    \label{eq:n}
    n_i = &-95.97 J_{i} + 25.54 J_{i}^2 -2.87 J_{i}^3\notag\\&+ 0.18 J_{i}^4 + 157.14
\end{align}
\subsection{Benchmark}
Our simulated hardware ELM is benchmarked on the MNIST dataset, as well as on two additional image datasets in order to assess its generalizability.
The first one is the EMNIST-letters dataset: an extension of MNIST to handwritten letters~\cite{cohen2017emnist}, consisting of $124,800$ training images and $20,800$ testing images classified into $26$ classes corresponding to the $26$ letters of the alphabet.
Each class contains lower and upper case letters. 
The second dataset is FMNIST~\cite{xiao2017fashion} and is composed of $60,000$ training images and $10,000$ testing images, all extracted from the Zalando website.
These images are separated into $10$ classes representing clothing items: \{\textit{t-shirt/top}, \textit{trouser}, \textit{pullover}, \textit{dress}, \textit{coat}, \textit{sandal}, \textit{shirt}, \textit{sneaker}, \textit{bag}, \textit{ankle boot}\}.

A STVO-based ELM is simulated and trained for each dataset accordingly to the procedure described in the previous section.
The number of PCA components selected is chosen to ensure an explained variance of $80\%$ in each training set, allowing to reduce the dimension of the data from $784$ to $38$ (resp. $24$) for the EMNIST-letters (resp. FMNIST) dataset.
For each dataset, we compute the test accuracy (\textit{i.e.}, the ratio of correctly classified unseen samples) and the normalized root-mean-square error (NRMSE) defined as in \eq\ref{eq:nrmse}~\cite{lukosevivcius2009reservoir}, with $\mathbf{\hat{y}}$ the output of the ELM, $\mathbf{t}$ the corresponding target, and $\langle\cdot\rangle$ the average operator.
\begin{equation}
    \label{eq:nrmse}
    \text{NRMSE} = \sqrt{\dfrac{\left\langle\vert\vert\mathbf{\hat{y}}-\mathbf{t}\vert\vert^2\right\rangle}{\left\langle\vert\vert\mathbf{t}-\left\langle\mathbf{t}\right\rangle\vert\vert^2\right\rangle}}
\end{equation}
The metrics are averaged over $10$ iterations using different randomly generated versions of $\mathbf{M}$, and the whole process is repeated for an increasing number of virtual neurons in the nonlinear layer (\textit{i.e.}, an increasing $N_\theta$).
This highlights the potential of DD-TEA for performing high-throughput parametric studies for hyperparameter optimization.

Finally, the performance are compared to the same architecture with more conventional nonlinear transfer functions such as the reLU and the sigmoid.
This is achieved by artificially replacing \eq(\ref{eq:hptea}) by the corresponding expressions in the simulations. 
The identity activation is also be tested as an activation function to underline the role of the nonlinearity in the classification ability of the ELM.\@

\section{Results}\label{sec:results}
\begin{figure}
    \centering
    \includegraphics[width=.8\columnwidth, trim={0cm 0cm 0cm 0cm}, clip]{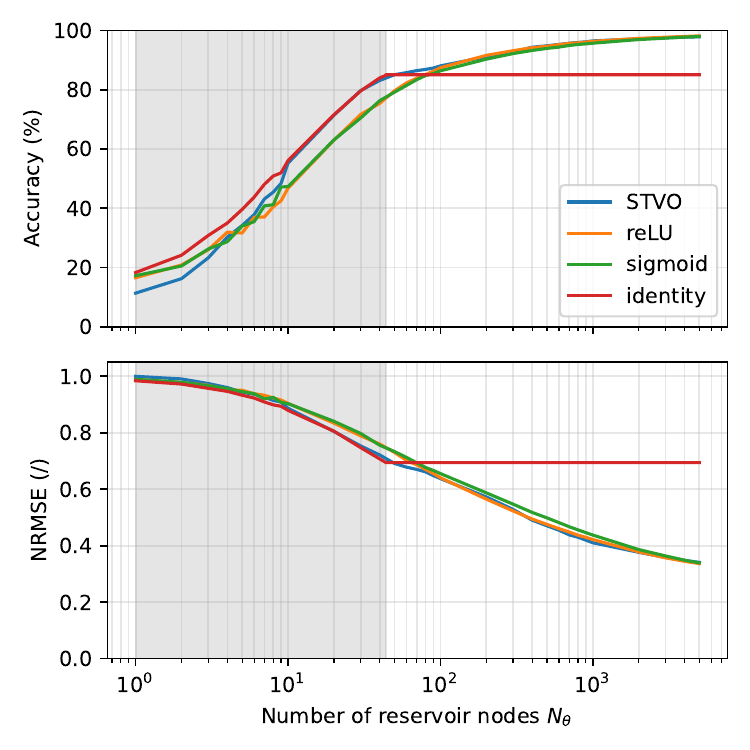}
    \caption{Test accuracy and NRMSE of our time-multiplexed ELM on the MNIST dataset for an increasing number of nodes.
    All reported accuracies are evaluated on the test set.
    The nonlinear transfer function of the reservoir nodes is switched between the STVO dynamics, the reLU, and the sigmoid functions.
    Linear nodes are also tested using the identity function.}\label{fig:overall}    
\end{figure}

The test accuracy and the NRMSE of our STVO-based ELM on the MNIST dataset are presented in \fig\ref{fig:overall}.
Without surprise, the accuracy (resp. NRMSE) increases (resp.\ decreases) with an increasing number of nodes $N_\theta$ in the nonlinear layer.
As a matter of fact, the number of nodes is directly proportional to the number of tunable parameters in the model (\textit{i.e.}, the number of elements in the learned matrix $\mathbf{W}$).
The shaded areas on the plots represent the range where the input data $\mathbf{x}$ is projected into a smaller dimension space due to the low amount of virtual nodes in the nonlinear layer (\textit{i.e.}, $N_\theta < N_f$), leading to data compression instead of expansion.
As a result, the performance is poor in this range, and similar to the one obtained with a linear classifier with $N_\theta$ parameters.
Yet, the performance still increase with $N_\theta$ in this regime, as it increases the size of the compression space and hence the quality of the linear classifier.
As soon as the size of the projection space reaches the dimension of the input data (\textit{i.e.}, $N_\theta\geq 44$), the performance obtained with the linear classifier saturates because the projected data simply become linear expansions of the input data, and the absence of nonlinearity prevents the learning of complex patterns (red curve in \fig\ref{fig:overall}).
More pragmatically, if we consider a linear transfer function represented by $s\left(\mathbf{x}''\right) = k\mathbf{x}''$ with $k$ a constant, the predicted category of an image $\mathbf{x}$ can be summarized by
\begin{equation}
    \label{eq:inf_summary}
    \hat{d} = \text{argmax}\left(\mathbf{T}_\text{train}\mathbf{S}^\dagger_\text{train}k\mathbf{M}\mathbf{C}\mathbf{x}\right).
\end{equation}
Substituting $\mathbf{S}_\text{train}$ by $k\mathbf{M}\mathbf{C}\mathbf{X}_{\text{train}}$, where $\mathbf{X}_{\text{train}}$ is the concatenation of the $60000$ training samples, yields
\begin{align}
    \label{eq:inf_summary2}
    \hat{d} &= \text{argmax}\left(\mathbf{T}_\text{train}{\left(k\mathbf{M}\mathbf{C}\mathbf{X}_{\text{train}}\right)}^\dagger k\mathbf{M}\mathbf{C}\mathbf{x}\right)\notag\\
    &= \text{argmax}\left(\mathbf{T}_\text{train}\mathbf{X}_{\text{train}}^\dagger {\left(k\mathbf{M}\mathbf{C}\right)}^\dagger k\mathbf{M}\mathbf{C}\mathbf{x}\right).
\end{align}
As ${\left(k\mathbf{M}\mathbf{C}\right)}^\dagger\left(k\mathbf{M}\mathbf{C}\right)=\mathbf{I}$ (the identity matrix), \eq(\ref{eq:inf_summary}) finally reduces to \eq(\ref{eq:linear}), which is independent from the network's parameters, and where the matrix $(\mathbf{T}_\text{train}\mathbf{X}_\text{train}^\dagger)$ contains the parameters of the linear regression between the training data and the training targets, applied to the input data $\mathbf{x}$.
\begin{equation}
    \label{eq:linear}
    \hat{d} = \text{argmax}\left(\mathbf{T}_\text{train}\mathbf{X}_{\text{train}}^\dagger\mathbf{x}\right)
\end{equation}
This is however not a case of overfitting.
Indeed, as the parameters are learned by linear regression over the whole training set, overfitting should not be observed as long as $N_\theta$ is lower than the number of training samples \textit{i.e.}, 60000.

\begin{figure}
    \centering
    \includegraphics[width=\columnwidth]{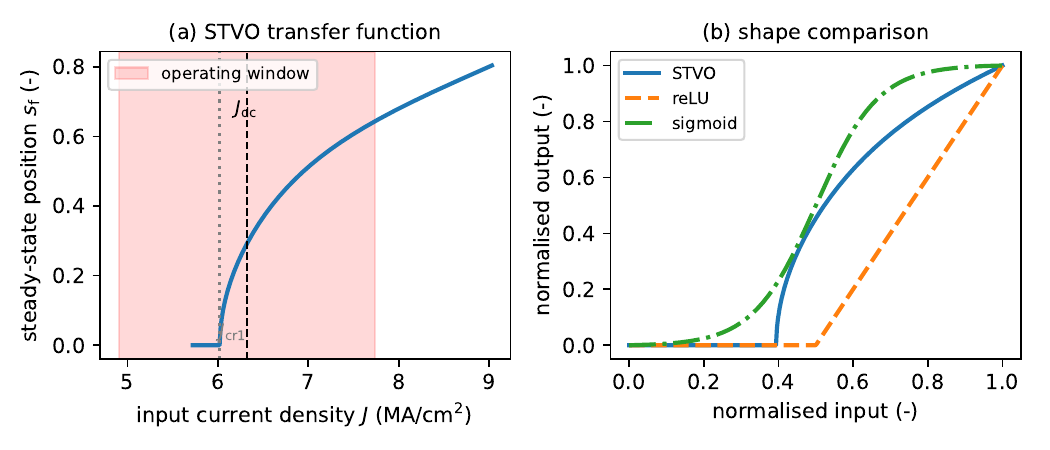}
    \caption{\textbf{(a)} Steady-state reduced core position $s_\text{f}$ as a function of the input current density $J$ for our STVO.\@ The shaded area is the operating window $[J_\text{dc}-\Delta J/2, J_\text{dc}+\Delta J/2]$, which contains the critical current density $J_\text{cr1}$.
    \textbf{(b)} Comparison of the normalized STVO transfer function with the reLU and sigmoid activations. The STVO nonlinearity interpolates between a rectifying and a saturating behavior.}\label{fig:activation}
\end{figure}

In all the nonlinear cases (STVO, reLU, and sigmoid), the performance improve steadily (accuracy increases and NRMSE decreases) with the number of parameters $N_\theta$. 
Eventually at higher $N_\theta$ values, the accuracy reaches a value common to all the activation functions, well above the saturation threshold reached with the identity activation, and similarly with the NRMSE.\@
The origin of this convergence becomes clear by looking at the effective nonlinearity implemented by the STVO dynamics.
Figure~\ref{fig:activation}a shows the steady-state reduced position $s_\text{f}(J)$ over the operating current window.
It can be seen that the STVO linearity acts as a smooth threshold-and-saturation function.
If normalized, its shape qualitatively interpolates between the rectifying reLU and the saturating sigmoid (\fig\ref{fig:activation}b), which explains why the three functions yield comparable random feature maps.
More fundamentally, a property of extreme learning machines is that any fixed nonlinearity provides a feature map rich enough for the linear readout as soon as $N_\theta$ is large enough to expand rather than compress the input, so the precise form of the activation does not matter anymore.
The presence of a large performance gap with respect to the identity activation for $N_\theta > 44$ confirms that some form of nonlinearity is however necessary.

The STVO dynamics also exhibit a short-term memory due to the transient coupling between successive virtual nodes, that the reLU and sigmoid do not possess.
For the image classification tasks considered in this work, this memory plays no role, and the three activations reach the same performance. 
For temporal tasks requiring fading memory like speech recognition or time series prediction, this equivalence would not hold anymore, and the STVO dynamics would reach higher performance than the reLU and the sigmoid.

\begin{figure}[h]
    \centering
    \includegraphics[width=.6\columnwidth]{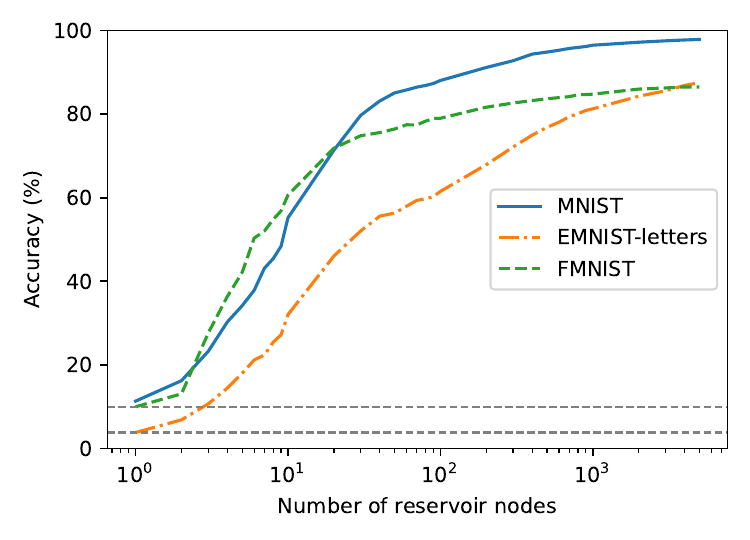}
    \caption{Test accuracy of the STVO-based ELM for the MNIST, EMNIST-letters and FMNIST datasets.
    The dashed lines represent the random choice accuracy levels linked to the number of categories in each dataset ($100\%/26 = 3.85\%$ and $100\%/10 = 10\%$).}\label{fig:accs}
\end{figure}

The comparison of the accuracy of the classification using the STVO-based ELM for the three datasets is represented in \fig\ref{fig:accs}.
It can be seen that for a very low number of virtual nodes in the reservoir, the accuracy reaches the random choice accuracy level \textit{i.e.}, $10\%$ for MNIST and FMNIST, and $3.85\%$ for EMNIST-letters.
The accuracy eventually reaches $97.8\%$ on MNIST and $87.5\%$ (resp. $86.5\%$) on the EMNIST-letters (resp. FMNIST) dataset with $5000$ nodes in the nonlinear layer.
The relatively low score achieved in FMNIST is due to the lack of convolutional layers in the architecture, something that is known to be of prior importance for the classification of complex images~\cite{lecun1989backpropagation}.
To test this hypothesis, we replaced the PCA preprocessing by a fixed random convolutional feature extractor (random filters followed by a reLU nonlinearity and max-pooling), reduced to the same dimensionality and fed to the identical ELM.\@
Even without any training of the convolutional filters, this consistently improves the FMNIST test accuracy (e.g.\ from $85.7\%$ to $86.2\%$ at $N_\theta = 1000$), confirming that convolutional processing is beneficial.
Trained convolutional layers would be expected to yield significantly larger gains.
Concerning EMNIST-letters, we suspect that the lower quality of the recognition is due to the presence of lower case and upper case characters in each category, which can be regarded as the presence of two distinct datasets with the same labels.
This assumption is consolidated by computing the average intraclass variance in each dataset.
The variance of each class $\mathbf{K}$ is computed accordingly to $\text{Var}_\text{intra}(\mathbf{K}) = \left(\sum_{i}(\mathbf{x}_i-\overline{\mathbf{K}})^2\right)/\vert\mathbf{K}\vert$ and averaged over all the classes for each dataset.
It is equal to $3401$ in FMNIST, $3452.9$ in MNIST, and $4408.7$ in EMNIST-letters, which is about $28\%$ larger than in the two former datasets.\@
The top accuracy reached on MNIST known to this day is $99.87\%$~\cite{byerly2021no}.
Our solution reaches an accuracy ($97.8\%$) competitive with other reservoir computing-based approaches in software and hardware~\cite{gardner2021modified, zhu2021mnist, gardner2022inkjet}, while relying on a much simpler architecture.
Other works indicate that higher performance could be achieved by increasing further the number of nodes in the reservoir~\cite{schaetti2016echo}.
The top accuracy on EMNIST-letters and FMNIST are respectively $95.96\%$ and $96.91\%$~\cite{jeevan2022wavemix, tanveer2021fine}.
However, these works are based on neural networks far more complex than ours, involving many specialized layers such as convolutional layers and attention modules, which strongly hinders their potential for hardware implementation and energy-efficient inference.

\section{Conclusion}
We designed an extreme learning machine based on non-linear spintronics oscillators called STVOs.
This approach aims at solving cognitive tasks more energy-efficiently than conventional software solutions by leveraging the low power consumption of STVOs as nonlinear data processors. 
The system has a simple architecture and only requires the learning of the readout weights through linear regression, while the rest of the parameters is random and fixed.
Moreover, the use of a single STVO delayed in time allows to further reduce the hardware complexity of the system, while still achieving state-of-the-art performance on the MNIST dataset.

To overcome the experimental load inherent to the optimization and testing of such a system, we used simulations of STVOs, allowing to obtain an explicit expression for the STVO dynamics that we used as neuronal nonlinearity.
This approach allowed to simulate an entire STVO-based ELM several orders of magnitude faster than previously available frameworks.
As DD-TEA has been quantitatively validated against micromagnetic simulations, the reference numerical method for STVO dynamics~\cite{araujo2022data,moureaux2023neuromorphic}, the present results rest on a physically grounded model.
Experimental validation on a physical device remains an essential next step for assessing the influence of device and measurement noise on the distribution of the STVO responses.
The simulation framework introduced here is precisely intended to prepare such experiments by narrowing the hyperparameter space beforehand.

Because the images are processed as static inputs, the system operates as an ELM rather than as a memory-dependent reservoir.
Therefore, the temporal multiplexing through a single STVO is used here as a hardware-efficient way to emulate a large fixed nonlinear feature map, not to exploit fading memory.
While other approaches treat each image as a genuine temporal signal~\cite{du2017reservoir} or rely on pretrained convolutional features within an ESN~\cite{ozdemir2022echovpr}, our static solution with PCA trades such temporal encoding and feature-extraction for simplicity.
In a physical realization, time-multiplexing $N_\theta$ virtual nodes requires injecting and sampling $N_\theta$ successive values through the single oscillator, \textit{i.e.} storing the masked sequence and spending a real time of $N_\theta \times D_\text{t}$ per inference.
For $N_\theta = 5000$ and $D_\text{t} = 10$~ns this results in $50$~\textmu s per image (without considering reset time and read-out overheads).
This constraint of single-node time-multiplexed hardware could be relaxed by spatially multiplexing several STVOs at the cost of additional chip area and increased complexity.

We showed that a STVO-based ELM is able to classify images of handwritten digits from the MNIST dataset with state-of-the-art accuracy levels.
The nonlinearity of STVO dynamics was found to be equivalent to more conventional nonlinear function such as the reLU and the sigmoid.
Moreover, the approach is easily generalizable to other datasets like the EMNIST-letters and the fashion MNIST datasets.
Better performance can likely be obtained on the two latter tasks by designing a more specialized architecture, for example involving convolutional data treatment before the reservoir~\cite{tong2018reservoir}. 

The use of the DD-TEA framework allowed to carry on extensive studies of the performance with respect to the number of reservoir nodes.
This could be reproduced with other hyperparameters such as the input and output shapes of the data or the properties of the input matrix $\mathbf{M}$. 
Physical operating parameters having an influence on the STVO dynamics could also be optimized the same way, allowing an efficient preparation of an experimental setup.  

Spin-torque oscillators have also been used for addressing temporal tasks such as spoken-digit recognition and chaotic time-series prediction~\cite{torrejon2017neuromorphic, riou2017neuromorphic, markovic2019reservoir, tsunegi2019physical}.
Extending the present DD-TEA framework to recurrent single-STVO reservoirs, in which the fading memory of the magnetization dynamics becomes a computational resource, is a natural next step.
More generally, we think that the DD-TEA framework could allow to develop and test other neural architectures based on STVOs to perform machine learning tasks with increased efficiency.
In parallel, increasing attention has been given to the matrix-vector multiplication step that allows to multiply the neurons ouputs by the weights of the following layer.
Recent works have demonstrated analog implementation of the so-called multiply-and-accumulate (MAC) operation using CMOS technology~\cite{paliy2020analog}, and others using the magnetic or RF properties of MTJs similar to the ones considered in this work~\cite{Jung2022, leroux2021hardware,leroux2021radio}.
This opens the path to the realization of hardware neuromorphic systems using MTJs for both nonlinear transformation and signal propagation though the layers, and hence to more energy-efficient machine learning solutions.

\section*{Data availability statement}
The data-driven STVO model, the extreme learning machine pipeline, and the scripts used to produce all the figures and reported metrics are available from the authors upon reasonable request.
The MNIST~\cite{deng2012mnist}, EMNIST-letters~\cite{cohen2017emnist} and Fashion MNIST~\cite{xiao2017fashion} datasets are publicly available.

\ack
Computational resources have been provided by the Consortium des Équipements de Calcul Intensif (CÉCI), funded by the Fonds de la Recherche Scientifique de Belgique (F.R.S.-FNRS) under Grant No. 2.5020.11 and by the Walloon Region.
F.A.A. is a Research Associate of the F.R.S.-FNRS.

\section*{References}
\bibliographystyle{iopart-num}
\bibliography{main}

@article{thiele1973steady,
  title     = {Steady-state motion of magnetic domains},
  author    = {Thiele, AA},
  journal   = {Physical Review Letters},
  volume    = {30},
  number    = {6},
  pages     = {230},
  year      = {1973},
  publisher = {APS},
  doi       = {10.1103/PhysRevLett.30.230}
}

@article{lecun1989backpropagation,
  title     = {Backpropagation applied to handwritten zip code recognition},
  author    = {LeCun, Yann and Boser, Bernhard and Denker, John S. and Henderson, Donnie and Howard, Richard E. and Hubbard, Wayne and Jackel, Lawrence D.},
  journal   = {Neural Computation},
  volume    = {1},
  number    = {4},
  pages     = {541--551},
  year      = {1989},
  publisher = {MIT Press},
  doi       = {10.1162/neco.1989.1.4.541}
}

@article{wilkes1995memory,
  title     = {The memory wall and the CMOS end-point},
  author    = {Wilkes, Maurice V},
  journal   = {ACM SIGARCH Computer Architecture News},
  volume    = {23},
  number    = {4},
  pages     = {4--6},
  year      = {1995},
  publisher = {ACM New York, NY, USA},
  doi       = {10.1145/218864.218865}
}

@article{frank2001device,
  title     = {Device scaling limits of Si MOSFETs and their application dependencies},
  author    = {Frank, David J and Dennard, Robert H and Nowak, Edward and Solomon, Paul M and Taur, Yuan and Wong, Hon-Sum Philip},
  journal   = {Proceedings of the IEEE},
  volume    = {89},
  number    = {3},
  pages     = {259--288},
  year      = {2001},
  publisher = {Ieee},
  doi       = {10.1109/5.915374}
}

@book{ben2003generalized,
  title={Generalized inverses: theory and applications},
  author={Ben-Israel, Adi and Greville, Thomas NE},
  volume={15},
  year={2003},
  publisher={Springer Science \& Business Media},
  doi = {10.1007/b97366}
}

@article{pribiag2007magnetic,
  title     = {Magnetic vortex oscillator driven by dc spin-polarized current},
  author    = {Pribiag, VS and Krivorotov, IN and Fuchs, GD and Braganca, PM and Ozatay, O and Sankey, JC and Ralph, DC and Buhrman, RA},
  journal   = {Nature Physics},
  volume    = {3},
  number    = {7},
  pages     = {498--503},
  year      = {2007},
  publisher = {Nature Publishing Group UK London},
  doi       = {10.1038/nphys619}
}

@article{yuasa2007giant,
  title     = {Giant tunnel magnetoresistance in magnetic tunnel junctions with a crystalline MgO (0 0 1) barrier},
  author    = {Yuasa, S and Djayaprawira, DD},
  journal   = {Journal of Physics D: Applied Physics},
  volume    = {40},
  number    = {21},
  pages     = {R337},
  year      = {2007},
  publisher = {IOP Publishing},
  doi       = {10.1088/0022-3727/40/21/R01}
}

@article{ralph2008spin,
  title     = {Spin transfer torques},
  author    = {Ralph, Daniel C and Stiles, Mark D},
  journal   = {Journal of Magnetism and Magnetic Materials},
  volume    = {320},
  number    = {7},
  pages     = {1190--1216},
  year      = {2008},
  publisher = {Elsevier},
  doi       = {10.1016/j.jmmm.2007.12.019}
}

@article{ringner2008principal,
  title     = {What is principal component analysis?},
  author    = {Ringn{\'e}r, Markus},
  journal   = {Nature Biotechnology},
  volume    = {26},
  number    = {3},
  pages     = {303--304},
  year      = {2008},
  publisher = {Nature Publishing Group US New York},
  doi       = {10.1038/nbt0308-303}
}

@article{courrieu2008fast,
  title   = {Fast computation of Moore-Penrose inverse matrices},
  author  = {Courrieu, Pierre},
  journal = {arXiv preprint arXiv:0804.4809},
  year    = {2008},
  doi     = {10.48550/arXiv.0804.4809}
}

@article{lukosevivcius2009reservoir,
  title     = {Reservoir computing approaches to recurrent neural network training},
  author    = {Luko{\v{s}}evi{\v{c}}ius, Mantas and Jaeger, Herbert},
  journal   = {Computer Science Review},
  volume    = {3},
  number    = {3},
  pages     = {127--149},
  year      = {2009},
  publisher = {Elsevier},
  doi       = {10.1016/j.cosrev.2009.03.005}
}

@article{gaididei2010magnetic,
  title     = {Magnetic vortex dynamics induced by an electrical current},
  author    = {Gaididei, Yuri and Kravchuk, Volodymyr P and Sheka, Denis D},
  journal   = {International Journal of Quantum Chemistry},
  volume    = {110},
  number    = {1},
  pages     = {83--97},
  year      = {2010},
  publisher = {Wiley Online Library},
  doi       = {10.1002/qua.22253}
}

@article{appeltant2011information,
  title     = {Information processing using a single dynamical node as complex system},
  author    = {Appeltant, Lennert and Soriano, Miguel Cornelles and Van der Sande, Guy and Danckaert, Jan and Massar, Serge and Dambre, Joni and Schrauwen, Benjamin and Mirasso, Claudio R and Fischer, Ingo},
  journal   = {Nature Communications},
  volume    = {2},
  number    = {1},
  pages     = {468},
  year      = {2011},
  publisher = {Nature Publishing Group UK London},
  doi       = {10.1038/ncomms1476}
}

@article{larger2012photonic,
  title     = {Photonic information processing beyond Turing: an optoelectronic implementation of reservoir computing},
  author    = {Larger, Laurent and Soriano, Miguel C and Brunner, Daniel and Appeltant, Lennert and Guti{\'e}rrez, Jose M and Pesquera, Luis and Mirasso, Claudio R and Fischer, Ingo},
  journal   = {Optics Express},
  volume    = {20},
  number    = {3},
  pages     = {3241--3249},
  year      = {2012},
  publisher = {Optical Society of America},
  doi       = {10.1364/OE.20.003241}
}

@article{paquot2012optoelectronic,
  title     = {Optoelectronic reservoir computing},
  author    = {Paquot, Yvan and Duport, Francois and Smerieri, Antoneo and Dambre, Joni and Schrauwen, Benjamin and Haelterman, Marc and Massar, Serge},
  journal   = {Scientific Reports},
  volume    = {2},
  number    = {1},
  pages     = {287},
  year      = {2012},
  publisher = {Nature Publishing Group UK London},
  doi       = {10.1038/srep00287}
}

@article{deng2012mnist,
  title     = {The MNIST database of handwritten digit images for machine learning research},
  author    = {Deng, Li},
  journal   = {IEEE Signal Processing Magazine},
  volume    = {29},
  number    = {6},
  pages     = {141--142},
  year      = {2012},
  publisher = {IEEE},
  doi       = {10.1109/MSP.2012.2211477}
}

@article{barata2012moore,
  title     = {The Moore-Penrose pseudoinverse: A tutorial review of the theory},
  author    = {Barata, Jo{\~a}o Carlos Alves and Hussein, Mahir Saleh},
  journal   = {Brazilian Journal of Physics},
  volume    = {42},
  pages     = {146--165},
  year      = {2012},
  publisher = {Springer},
  doi       = {10.1007/s13538-011-0052-z}
}

@article{karamizadeh2013overview,
  title     = {An overview of principal component analysis},
  author    = {Karamizadeh, Sasan and Abdullah, Shahidan M and Manaf, Azizah A and Zamani, Mazdak and Hooman, Alireza},
  journal   = {Journal of Signal and Information Processing},
  volume    = {4},
  number    = {3B},
  pages     = {173},
  year      = {2013},
  publisher = {Scientific Research Publishing},
  doi       = {10.4236/jsip.2013.43B031}
}

@article{guslienko2014nonlinear,
  title     = {Nonlinear magnetic vortex dynamics in a circular nanodot excited by spin-polarized current},
  author    = {Guslienko, Konstantin Y and Sukhostavets, Oksana V and Berkov, Dmitry V},
  journal   = {Nanoscale Research Letters},
  volume    = {9},
  pages     = {1--7},
  year      = {2014},
  publisher = {Springer},
  doi       = {10.1186/1556-276X-9-386}
}

@article{ortin2015unified,
  title     = {A unified framework for reservoir computing and extreme learning machines based on a single time-delayed neuron},
  author    = {Ort{\'\i}n, S and Soriano, Miguel C and Pesquera, L and Brunner, Daniel and San-Mart{\'\i}n, D and Fischer, Ingo and Mirasso, CR and Guti{\'e}rrez, JM},
  journal   = {Scientific Reports},
  volume    = {5},
  number    = {1},
  pages     = {14945},
  year      = {2015},
  publisher = {Nature Publishing Group UK London},
  doi       = {10.1038/srep14945}
}

@article{yogendra2015coupled,
  title     = {Coupled Spin Torque Nano Oscillators for Low Power Neural Computation},
  author    = {Yogendra, Karthik and Fan, Deliang and Roy, Kaushik},
  journal   = {IEEE Transactions on Magnetics},
  volume    = {51},
  number    = {10},
  pages     = {1--9},
  year      = {2015},
  publisher = {IEEE},
  doi       = {10.1109/TMAG.2015.2443042}
}

@inproceedings{schaetti2016echo,
  title        = {Echo state networks-based reservoir computing for mnist handwritten digits recognition},
  author       = {Schaetti, Nils and Salomon, Michel and Couturier, Rapha{\"e}l},
  booktitle    = {2016 IEEE Intl Conference on Computational Science and Engineering (CSE) and IEEE Intl Conference on Embedded and Ubiquitous Computing (EUC) and 15th Intl Symposium on Distributed Computing and Applications for Business Engineering (DCABES)},
  pages        = {484--491},
  year         = {2016},
  organization = {IEEE},
  doi          = {10.1109/CSE-EUC-DCABES.2016.229}
}

@article{larger2017high,
  title     = {High-speed photonic reservoir computing using a time-delay-based architecture: Million words per second classification},
  author    = {Larger, Laurent and Bayl{\'o}n-Fuentes, Antonio and Martinenghi, Romain and Udaltsov, Vladimir S and Chembo, Yanne K and Jacquot, Maxime},
  journal   = {Physical Review X},
  volume    = {7},
  number    = {1},
  pages     = {011015},
  year      = {2017},
  publisher = {APS},
  doi       = {10.1103/PhysRevX.7.011015}
}

@inproceedings{riou2017neuromorphic,
  title        = {Neuromorphic computing through time-multiplexing with a spin-torque nano-oscillator},
  author       = {Riou, Mathieu and Abreu Araujo, Flavio and Torrejon, Jacob and Tsunegi, Sumito and Khalsa, Guru and Querlioz, Damien and Bortolotti, Paolo and Cros, Vincent and Yakushiji, Kay and Fukushima, Akio and others},
  booktitle    = {2017 IEEE International Electron Devices Meeting (IEDM)},
  pages        = {36--3},
  year         = {2017},
  organization = {IEEE},
  doi          = {10.1109/IEDM.2017.8268505}
}

@article{torrejon2017neuromorphic,
  title     = {Neuromorphic computing with nanoscale spintronic oscillators},
  author    = {Torrejon, Jacob and Riou, Mathieu and Abreu Araujo, Flavio and Tsunegi, Sumito and Khalsa, Guru and Querlioz, Damien and Bortolotti, Paolo and Cros, Vincent and Yakushiji, Kay and Fukushima, Akio and others},
  journal   = {Nature},
  volume    = {547},
  number    = {7664},
  pages     = {428--431},
  year      = {2017},
  publisher = {Nature Publishing Group UK London},
  doi       = {10.1038/nature23011}
}

@inproceedings{cohen2017emnist,
  title        = {EMNIST: Extending MNIST to handwritten letters},
  author       = {Cohen, Gregory and Afshar, Saeed and Tapson, Jonathan and Van Schaik, Andre},
  booktitle    = {2017 International Joint Conference on Neural Networks (IJCNN)},
  pages        = {2921--2926},
  year         = {2017},
  organization = {IEEE},
  doi          = {10.48550/arXiv.1702.05373}
}

@article{xiao2017fashion,
  title   = {Fashion-mnist: a novel image dataset for benchmarking machine learning algorithms},
  author  = {Xiao, Han and Rasul, Kashif and Vollgraf, Roland},
  journal = {arXiv preprint arXiv:1708.07747},
  year    = {2017},
  doi     = {10.48550/arXiv.1708.07747}
}

@article{romera2018vowel,
  title     = {Vowel recognition with four coupled spin-torque nano-oscillators},
  author    = {Romera, Miguel and Talatchian, Philippe and Tsunegi, Sumito and Abreu Araujo, Flavio and Cros, Vincent and Bortolotti, Paolo and Trastoy, Juan and Yakushiji, Kay and Fukushima, Akio and Kubota, Hitoshi and others},
  journal   = {Nature},
  volume    = {563},
  number    = {7730},
  pages     = {230--234},
  year      = {2018},
  publisher = {Nature Publishing Group UK London},
  doi       = {10.1038/s41586-018-0632-y}
}

@inproceedings{tong2018reservoir,
  title        = {Reservoir computing with untrained convolutional neural networks for image recognition},
  author       = {Tong, Zhiqiang and Tanaka, Gouhei},
  booktitle    = {2018 24th International Conference on Pattern Recognition (ICPR)},
  pages        = {1289--1294},
  year         = {2018},
  organization = {IEEE},
  doi          = {10.1109/ICPR.2018.8545471}
}

@article{tanaka2019recent,
  title     = {Recent advances in physical reservoir computing: A review},
  author    = {Tanaka, Gouhei and Yamane, Toshiyuki and H{\'e}roux, Jean Benoit and Nakane, Ryosho and Kanazawa, Naoki and Takeda, Seiji and Numata, Hidetoshi and Nakano, Daiju and Hirose, Akira},
  journal   = {Neural Networks},
  volume    = {115},
  pages     = {100--123},
  year      = {2019},
  publisher = {Elsevier},
  doi       = {10.1016/j.neunet.2019.03.005}
}

@article{markovic2019reservoir,
  title     = {Reservoir computing with the frequency, phase, and amplitude of spin-torque nano-oscillators},
  author    = {Markovi{\'c}, Danijela and Leroux, Nathan and Riou, Mathieu and Abreu Araujo, Flavio and Torrejon, Jacob and Querlioz, Damien and Fukushima, Akio and Yuasa, Shinji and Trastoy, Juan and Bortolotti, Paolo and others},
  journal   = {Applied Physics Letters},
  volume    = {114},
  number    = {1},
  year      = {2019},
  publisher = {AIP Publishing},
  doi       = {10.1063/1.5079305}
}

@article{leliaert2019tomorrow,
  title     = {Tomorrow’s micromagnetic simulations},
  author    = {Leliaert, Jonathan and Mulkers, Jeroen},
  journal   = {Journal of Applied Physics},
  volume    = {125},
  number    = {18},
  year      = {2019},
  publisher = {AIP Publishing},
  doi       = {10.1063/1.5093730}
}

@article{abreu2020role,
  title     = {Role of non-linear data processing on speech recognition task in the framework of reservoir computing},
  author    = {Abreu Araujo, Flavio and Riou, Mathieu and Torrejon, Jacob and Tsunegi, Sumito and Querlioz, Damien and Yakushiji, Kay and Fukushima, Akio and Kubota, Hitoshi and Yuasa, Shinji and Stiles, Mark D and others},
  journal   = {Scientific Reports},
  volume    = {10},
  number    = {1},
  pages     = {328},
  year      = {2020},
  publisher = {Nature Publishing Group UK London},
  doi       = {10.1038/s41598-019-56991-x}
}

@article{dhar2020carbon,
  title   = {The carbon impact of artificial intelligence},
  author  = {Dhar, Payal},
  journal = {Nature Machine Intelligence},
  volume  = {2},
  number  = {8},
  pages   = {423--425},
  year    = {2020},
  doi     = {10.1038/s42256-020-0219-9}
}

@article{paliy2020analog,
  title={Analog vector-matrix multiplier based on programmable current mirrors for neural network integrated circuits},
  author={Paliy, Maksym and Strangio, Sebastiano and Ruiu, Piero and Rizzo, Tommaso and Iannaccone, Giuseppe},
  journal={IEEE Access},
  volume={8},
  pages={203525--203537},
  year={2020},
  publisher={IEEE},
  doi={10.1109/ACCESS.2020.3037017}
}

@article{borghi2021reservoir,
  title     = {Reservoir computing based on a silicon microring and time multiplexing for binary and analog operations},
  author    = {Borghi, Massimo and Biasi, Stefano and Pavesi, Lorenzo},
  journal   = {Scientific Reports},
  volume    = {11},
  number    = {1},
  pages     = {15642},
  year      = {2021},
  publisher = {Nature Publishing Group UK London},
  doi       = {10.1038/s41598-021-94952-5}
}

@article{byerly2021no,
  title     = {No routing needed between capsules},
  author    = {Byerly, Adam and Kalganova, Tatiana and Dear, Ian},
  journal   = {Neurocomputing},
  volume    = {463},
  pages     = {545--553},
  year      = {2021},
  publisher = {Elsevier},
  doi       = {10.48550/arXiv.2001.09136}
}

@inproceedings{gardner2021modified,
  title        = {A modified echo state network for time independent image classification},
  author       = {Gardner, Steven D and Haider, Mohammad R and Moradi, Lee and Vantsevich, Vladimir},
  booktitle    = {2021 IEEE International Midwest Symposium on Circuits and Systems (MWSCAS)},
  pages        = {255--258},
  year         = {2021},
  organization = {IEEE},
  doi          = {10.1109/MWSCAS47672.2021.9531776}
}

@inproceedings{zhu2021mnist,
  title     = {MNIST classification using neuromorphic nanowire networks},
  author    = {Zhu, Ruomin and Loeffler, Alon and Hochstetter, Joel and Diaz-Alvarez, Adrian and Nakayama, Tomonobu and Stieg, Adam and Gimzewski, James and Lizier, Joseph and Kuncic, Zdenka},
  booktitle = {International Conference on Neuromorphic Systems 2021},
  pages     = {1--4},
  year      = {2021},
  doi       = {10.1145/3477145.3477162}
}

@inproceedings{tanveer2021fine,
  title        = {Fine-tuning darts for image classification},
  author       = {Tanveer, Muhammad Suhaib and Khan, Muhammad Umar Karim and Kyung, Chong-Min},
  booktitle    = {2020 25th International Conference on Pattern Recognition (ICPR)},
  pages        = {4789--4796},
  year         = {2021},
  organization = {IEEE},
  doi          = {10.48550/arXiv.2006.09042}
}

@article{leroux2021hardware,
  title={Hardware realization of the multiply and accumulate operation on radio-frequency signals with magnetic tunnel junctions},
  author={Leroux, Nathan and Mizrahi, Alice and Markovi{\'c}, Danijela and Sanz-Hern{\'a}ndez, D{\'e}dalo and Trastoy, Juan and Bortolotti, Paolo and Martins, Leandro and Jenkins, Alex and Ferreira, Ricardo and Grollier, Julie},
  journal={Neuromorphic Computing and Engineering},
  volume={1},
  number={1},
  pages={011001},
  year={2021},
  publisher={IOP Publishing},
  doi={10.1088/2634-4386/abfca6}
}

@article{leroux2021radio,
  title={Radio-frequency multiply-and-accumulate operations with spintronic synapses},
  author={Leroux, Nathan and Markovi{\'c}, Danijela and Martin, Erwann and Petrisor, Teodora and Querlioz, Damien and Mizrahi, Alice and Grollier, Julie},
  journal={Physical Review Applied},
  volume={15},
  number={3},
  pages={034067},
  year={2021},
  publisher={APS},
  doi={10.1103/PhysRevApplied.15.034067}
}

@article{araujo2022data,
  title   = {Data-driven Thiele equation approach for solving the full nonlinear spin-torque vortex oscillator dynamics},
  author  = {Abreu Araujo, Flavio and Chopin, Chlo{\'e} and de Wergifosse, Simon},
  journal = {arXiv preprint arXiv:2206.13596},
  year    = {2022},
  doi     = {10.48550/arXiv.2206.13596}
}

@article{abreu2022ampere,
  title     = {Ampere-Oersted field splitting of the nonlinear spin-torque vortex oscillator dynamics},
  author    = {Abreu Araujo, Flavio and Chopin, Chlo{\'e} and de Wergifosse, Simon},
  journal   = {Scientific Reports},
  volume    = {12},
  number    = {1},
  pages     = {10605},
  year      = {2022},
  publisher = {Nature Publishing Group UK London},
  doi       = {10.1038/s41598-022-14574-3}
}

@article{gardner2022inkjet,
  title     = {An inkjet-printed artificial neuron for physical reservoir computing},
  author    = {Gardner, Steven D and Haider, Mohammad R},
  journal   = {IEEE Journal on Flexible Electronics},
  volume    = {1},
  number    = {3},
  pages     = {185--193},
  year      = {2022},
  publisher = {IEEE},
  doi       = {10.1109/JFLEX.2022.3193346}
}

@article{jeevan2022wavemix,
  title   = {Wavemix-lite: A resource-efficient neural network for image analysis},
  author  = {Jeevan, Pranav and Viswanathan, Kavitha and Sethi, Amit},
  journal = {arXiv preprint arXiv:2205.14375},
  year    = {2022},
  doi     = {10.48550/arXiv.2205.14375}
}

@article{Jung2022,
    author = {Jung, S. and Lee, H. and Myung, S. and Kim, H. and Yoon, S. K. and Kwon, S.-W. and Ju, Y. and Kim, M. and Yi, W. and Han, S. and Kwon, B. and Seo, B. and Lee, K. and Koh, G.-H. and Lee, K. and Song, Y. and Choi, C. and Ham, D and Kim, S. J.},
    title = {A crossbar array of magnetoresistive memory devices for in-memory computing},
    journal = {Nature},
    volume = {2},
    number = {9},
    pages = {211-216},
    year = {2022},
    doi = {https://doi.org/10.1038/s41586-021-04196-6}
}

@article{moureaux2023neuromorphic,
  title   = {Neuromorphic spintronics accelerated by an unconventional data-driven Thiele equation approach},
  author  = {Moureaux, Anatole and de Wergifosse, Simon and Chopin, Chlo{\'e} and Weber, Jimmy and Abreu Araujo, Flavio},
  journal = {arXiv preprint arXiv:2301.11025},
  year    = {2023},
  doi     = {10.48550/arXiv.2301.11025}
}

@article{grollier2020neuromorphic,
  title   = {Neuromorphic spintronics},
  author  = {Grollier, Julie and Querlioz, Damien and Camsari, K. Y. and Everschor-Sitte, Karin and Fukami, Shunsuke and Stiles, Mark D.},
  journal = {Nature Electronics},
  volume  = {3},
  number  = {7},
  pages   = {360--370},
  year    = {2020},
  doi     = {10.1038/s41928-019-0360-9}
}

@article{tsunegi2019physical,
  title   = {Physical reservoir computing based on spin torque oscillator with forced synchronization},
  author  = {Tsunegi, Sumito and Taniguchi, Tomohiro and Nakajima, Kohei and Miwa, Shinji and Yakushiji, Kay and Fukushima, Akio and Yuasa, Shinji and Kubota, Hitoshi},
  journal = {Applied Physics Letters},
  volume  = {114},
  number  = {16},
  pages   = {164101},
  year    = {2019},
  doi     = {10.1063/1.5081797}
}

@article{du2017reservoir,
  title   = {Reservoir computing using dynamic memristors for temporal information processing},
  author  = {Du, Chao and Cai, Fuxi and Zidan, Mohammed A. and Ma, Wen and Lee, Seung Hwan and Lu, Wei D.},
  journal = {Nature Communications},
  volume  = {8},
  number  = {1},
  pages   = {2204},
  year    = {2017},
  doi     = {10.1038/s41467-017-02337-y}
}

@article{ozdemir2022echovpr,
  title   = {{EchoVPR}: Echo State Networks for Visual Place Recognition},
  author  = {Ozdemir, Anil and Scerri, Mark and Barron, Andrew B. and Philippides, Andrew and Mangan, Michael and Vasilaki, Eleni and Manneschi, Luca},
  journal = {IEEE Robotics and Automation Letters},
  volume  = {7},
  number  = {2},
  pages   = {4520--4527},
  year    = {2022},
  doi     = {10.1109/LRA.2022.3150505}
}

@article{lecun2015deep,
  title   = {Deep learning},
  author  = {LeCun, Yann and Bengio, Yoshua and Hinton, Geoffrey},
  journal = {Nature},
  volume  = {521},
  number  = {7553},
  pages   = {436--444},
  year    = {2015},
  doi     = {10.1038/nature14539}
}

@article{patterson2021carbon,
  title   = {Carbon emissions and large neural network training},
  author  = {Patterson, David and Gonzalez, Joseph and Le, Quoc and Liang, Chen and Munguia, Lluis-Miquel and Rothchild, Daniel and So, David and Texier, Maud and Dean, Jeff},
  journal = {arXiv preprint arXiv:2104.10350},
  year    = {2021},
  doi     = {10.48550/arXiv.2104.10350}
}

@article{devries2023growing,
  title   = {The growing energy footprint of artificial intelligence},
  author  = {de Vries, Alex},
  journal = {Joule},
  volume  = {7},
  number  = {10},
  pages   = {2191--2194},
  year    = {2023},
  doi     = {10.1016/j.joule.2023.09.004}
}

@article{waldrop2016chips,
  title   = {The chips are down for Moore's law},
  author  = {Waldrop, M. Mitchell},
  journal = {Nature},
  volume  = {530},
  number  = {7589},
  pages   = {144--147},
  year    = {2016},
  doi     = {10.1038/530144a}
}

@article{markovic2020physics,
  title   = {Physics for neuromorphic computing},
  author  = {Markovi{\'c}, Danijela and Mizrahi, Alice and Querlioz, Damien and Grollier, Julie},
  journal = {Nature Reviews Physics},
  volume  = {2},
  number  = {9},
  pages   = {499--510},
  year    = {2020},
  doi     = {10.1038/s42254-020-0208-2}
}

@article{jaeger2004harnessing,
  title   = {Harnessing nonlinearity: Predicting chaotic systems and saving energy in wireless communication},
  author  = {Jaeger, Herbert and Haas, Harald},
  journal = {Science},
  volume  = {304},
  number  = {5667},
  pages   = {78--80},
  year    = {2004},
  doi     = {10.1126/science.1091277}
}

@article{huang2006extreme,
  title   = {Extreme learning machine: Theory and applications},
  author  = {Huang, Guang-Bin and Zhu, Qin-Yu and Siew, Chee-Kheong},
  journal = {Neurocomputing},
  volume  = {70},
  number  = {1-3},
  pages   = {489--501},
  year    = {2006},
  doi     = {10.1016/j.neucom.2005.12.126}
}
\end{document}